\documentclass[conference]{IEEEtran}
\IEEEoverridecommandlockouts
\usepackage{cite}
\usepackage{graphicx}
\usepackage{amsmath,amssymb,amsfonts}
\usepackage{tabularx}
\usepackage{booktabs}
\usepackage{algorithmic}
\usepackage{graphicx}
\usepackage{textcomp}
\usepackage{xcolor}
\def\BibTeX{{\rm B\kern-.05em{\sc i\kern-.025em b}\kern-.08em
    T\kern-.1667em\lower.7ex\hbox{E}\kern-.125emX}}
\begin{document}

\title{ReActXGB: A Hybrid Binary Convolutional Neural Network Architecture for Improved Performance and Computational Efficiency\\

}

\author{\IEEEauthorblockN{Po-Hsun Chu}
\IEEEauthorblockA{\textit{Depart. of Computer Science and Information Engineering} \\
\textit{National Central University}\\
Taoyuan, Taiwan \\
benny890808@gmail.com}
\and
\IEEEauthorblockN{Ching-Han Chen\IEEEauthorrefmark{1}\thanks{* Corresponding author}}
\IEEEauthorblockA{\textit{Depart. of Computer Science and Information Engineering} \\
\textit{National Central University}\\
Taoyuan, Taiwan \\
pierre@g.ncu.edu.tw}
}

\maketitle

\begin{abstract}
Binary convolutional neural networks (BCNNs) provide a potential solution to reduce the memory requirements and computational costs associated with deep neural networks (DNNs). However,  achieving a trade-off between performance and computational resources remains a significant challenge. Furthermore, the fully connected layer of BCNNs has evolved into a significant computational bottleneck. This is mainly due to the conventional practice of excluding the input layer and fully connected layer from binarization to prevent a substantial loss in accuracy. In this paper, we propose a hybrid model named ReActXGB, where we replace the fully convolutional layer of ReActNet-A with XGBoost. This modification targets to narrow the performance gap between BCNNs and real-valued networks while maintaining lower computational costs. Experimental results on the FashionMNIST benchmark demonstrate that ReActXGB outperforms ReActNet-A by 1.47\% in top-1 accuracy, along with a reduction of 7.14\% in floating-point operations (FLOPs) and 1.02\% in model size.
\end{abstract}

\vspace{\baselineskip}

\section{Introduction}

Binary convolutional neural networks (BCNNs) revolutionize neural networks, particularly in deep learning, through radical quantization known as binarization. In BCNNs, weights and activations in convolutional neural networks (CNNs) are simplified to a single bit, diverging from traditional real-valued representations. This significant precision reduction caters to the demand for efficient, resource-friendly models, especially for deployment on edge devices like wearables and tiny sensors.

Despite the hardware-friendly features, BCNNs face a significant challenge: a substantial decrease in accuracy. Consequently, BCNNs like ReActNet-A \cite{b1} and Sparks \cite{b2} preserve full precision in input and fully connected layers to maintain model generalization. While some BCNNs, e.g., FracBNN \cite{b3}, optimize the input layer, none address the fully connected layer, making it a bottleneck in BCNN computations.


In this study, considering ReActNet-A \cite{b1} as the state-of-the-art BCNN on ImageNet and recognizing XGBoost \cite{b4} for its representation in machine learning, along with its short computation times and minimal hardware resource consumption. Moreover, XGBoost \cite{b4}, being composed of multiple decision trees, enables efficient parallel processing in hardware implementation. Consequently, we have opted to deploy ReActNet-A \cite{b1} as the feature extractor and leverage XGBoost \cite{b4} as the classifier. This model, dubbed ReActXGB, aims to significantly reduce computational costs and memory size on the fully connected layer.

\section{Method}
\begin{figure}[htbp]
\centerline{\includegraphics[width=1\linewidth]{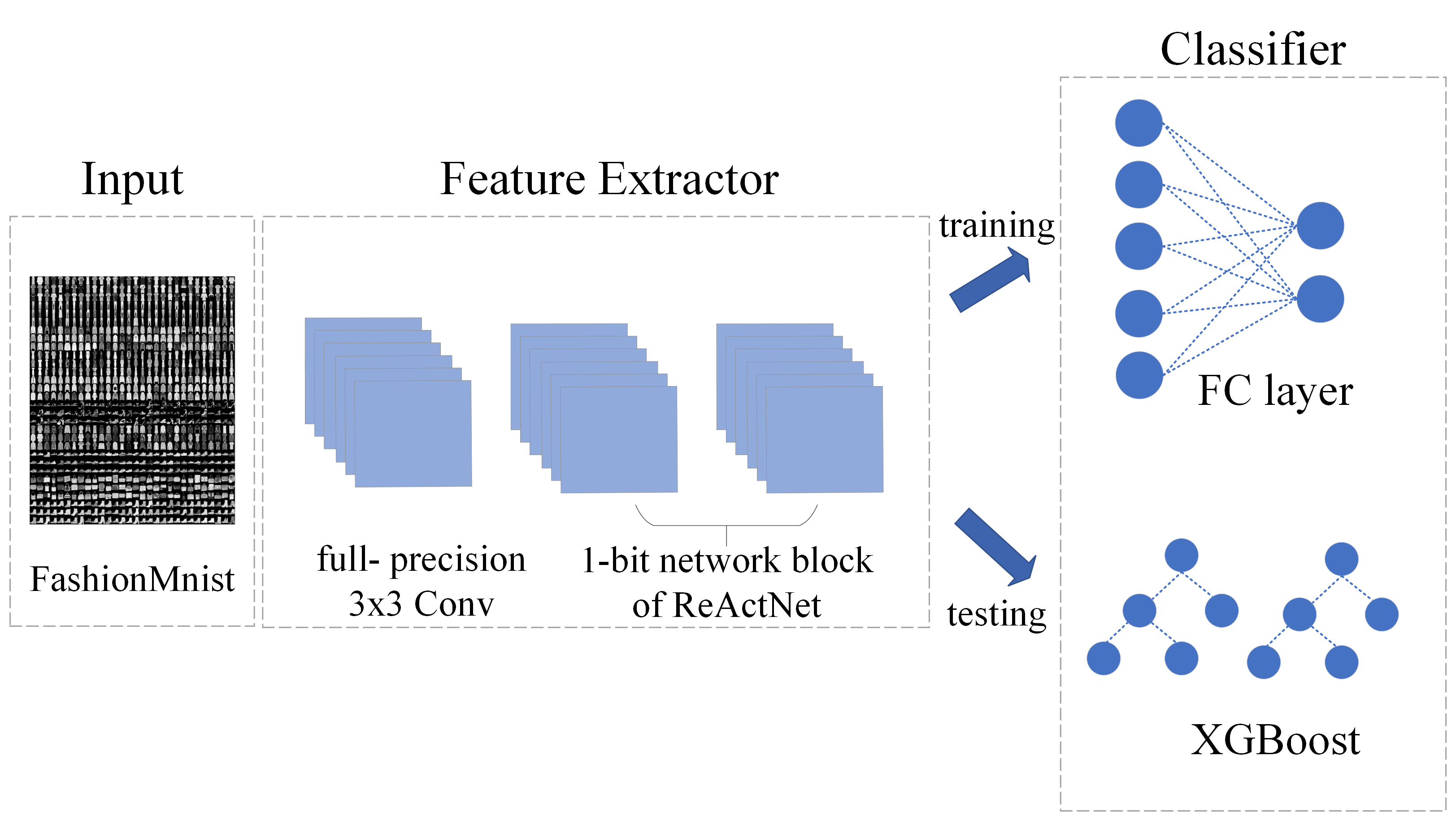}}
\caption{ReActXGB model architecture}
\label{fig1}
\end{figure}

This section will provide a detailed discussion of the model architecture, as shown in Fig.~\ref{fig1}, divided into three parts. The first part focuses on feature extraction, the second part on classification, and the third part on the integration of these two models.

\subsection{Feature extraction design}

We employ a 32-bit 3x3 convolution as our first layer, and utilize ReActNet-A's \cite{b1} 1-bit network block for our subsequent convolutional layers. The results of the convolution operation serve as input to the classifier.

\subsection{Classifier design}

Due to the ensemble decision tree nature of XGBoost \cite{b4}, which differs from BCNN in terms of training methodology, we adopt a two-stage training approach. In the first stage, we use a fully connected layer as the classifier for training model parameters, employing a stochastic gradient descent optimizer for parameter updates. In the second stage, we utilize the features outputted by the last convolutional layer as input for training XGBoost \cite{b4}. During training, we limit XGBoost \cite{b4} to be composed of a maximum of 20 decision trees, each with a maximum depth of 10. This not only maintains accuracy but also constrains hardware resource consumption. In the inference stage, an end-to-end approach successfully addresses the challenges of high parameter count and computational complexity associated with fully connected layers.

\section{Experiment Result}

We benchmark our model using FashionMNIST against baseline models, ResNet-18 and ReActNet-A. Our evaluation encompasses three key aspects: accuracy, computation cost, and parameter count.

\begin{table}[htb]
\centering
\caption{The top-1 accuracy of the models on FashionMNIST}
\begin{tabular}{|c|c|}
\hline   \textbf{Model} & \textbf{Top-1 acc (\%)} \\   
\hline   ResNet-18 \cite{b5} &  92.18 \\ 
\hline   ReActNet-A \cite{b1} &  88.91 \\  
\hline   \textbf{our proposed} & \textbf{90.38}  \\
\hline   
\end{tabular}
\label{t1}
\end{table}

\begin{table}[htb]
\centering
\caption{The computational cost of the models on FashionMNIST}
\begin{tabular}{|c|c|c|c|}
\hline   \textbf{Model} & \textbf{BOPs (x$10^{\mathrm{8}}$)} & \textbf{FLOPs (x$10^{\mathrm{6}}$)} &  \textbf{OPs (x$10^{\mathrm{6}}$) \cite{b6}}\\   
\hline   ResNet-18 \cite{b5} &  0 & 33 & 33 \\ 
\hline   ReActNet-A \cite{b1} &  1.38 & 0.14 & 2.3 \\  
\hline   \textbf{our proposed} & \textbf{1.38} & \textbf{0.13} & \textbf{2.29}  \\
\hline  
\multicolumn{4}{l}{\textsuperscript{\textit{} \(\text{OPs} = \frac{\text{BOPs}}{64} + \text{FLOPs}\).}}
\end{tabular}
\label{t2}
\end{table}

\begin{table}[htb]
\centering
\caption{The parameter of the models on FashionMNIST}
\begin{tabular}{|c|c|}
\hline   \textbf{Model} & \textbf{Parameter (MB)} \\   
\hline   ResNet-18 \cite{b5} &  42.67 \\ 
\hline   ReActNet-A \cite{b1} &  3.91 \\  
\hline   \textbf{our proposed} & \textbf{3.87}  \\
\hline   
\end{tabular}
\label{t3}
\end{table}

The limited deployability of BCNN on edge devices has been a persistent challenge, primarily due to its lower recognition accuracy. However, as indicated in TABLE~\ref{t1}, we observe that our model achieves an Top-1 accuracy 1.47\% higher than ReActNet-A \cite{b1}. Furthermore, when comparing ReActNet-A \cite{b1} and our model against ResNet-18 \cite{b5} simultaneously, our approach demonstrates a reduction in the performance gap between BCNN and modern CNNs.

In addition, while our model improves accuracy, it significantly reduces FLOPs computations by 7.14\% compared to ReActNet-A \cite{b1}. Consequently, OPs also decreases by 0.4\%, as shown in TABLE~\ref{t2}. The formula of OPs is calculated by OPs = BOPs/64 + FLOPs, following [6].

Regarding parameters, considering the inclusion of both 32-bit and 1-bit operations in the model structure, representing the computation in MetaBytes provides a more practical reflection of the hardware workload. As shown in TABLE~\ref{t3}, our model reduces parameter count by 1.02\% compared to ReActNet-A \cite{b1}.

With a substantial reduction in both parameter count and model computational complexity, coupled with an improvement in accuracy, even competing closely with ResNet-18 \cite{b5} with an accuracy difference of less than 2\%, it is evident that this approach has significant implications for the field of model compression research.

\section{Conclusion}

By leveraging the outstanding classification performance and simple computational complexity exhibited by XGBoost \cite{b4}, ReActXGB effectively addresses challenges associated with high computational load and parameter abundance in traditional fully connected layers. Integrating XGBoost \cite{b4} as a replacement and utilizing ReActNet-A \cite{b1} as a feature extractor, our model reduces parameter count and computational complexity while improving accuracy. This model shows promise for hardware implementation, particularly in electronic products requiring real-time processing or energy efficiency. In future work, we aim to implement ReActXGB on hardware, potentially using FPGA for an AI hardware accelerator, allowing us to measure and evaluate its real-world performance, power consumption, and hardware resource utilization, offering valuable insights for practical deployment in hardware applications.

\section*{Acknowledgment}

The authors would like to thank the National Science and Technology Council, Taiwan, R.O.C., for their financial support (Grant Nos. NSTC 112-2813-C-008 -040 -E and NSTC 112-2221-E-008 -029 -MY3)

\end{document}